# DDI-100: Dataset for Text Detection and Recognition

*Ilia* Zharikov[1,*], *Filipp* Nikitin[1], *Ilia* Vasiliev[1], and *Vladimir* Dokholyan[1]

[1]Moscow Institute of Physics and Technology

**Abstract.** Nowadays document analysis and recognition remain challenging tasks. However, only a few datasets designed for text detection (TD) and optical character recognition (OCR) problems exist. In this paper we present Distorted Document Images dataset (DDI-100) and demonstrate its usefulness in a wide range of document analysis problems. DDI-100 dataset is a synthetic dataset based on 7000 real unique document pages and consists of more than 100000 augmented images. Ground truth comprises text and stamp masks, text and characters bounding boxes with relevant annotations. Validation of DDI-100 dataset was conducted using several TD and OCR models that show high-quality performance on real data.

## 1 Introduction

Nowadays document image analysis is still a relevant and challenging problem in computer vision [1-4]. The ability to recognize document by its photo can simplify the process of document flow and help with numerous real-world tasks, for example, text detection and recognition [5], document image dewarping [6-7], layout recognition [8], etc. However, according to our best knowledge, all publicly available datasets for most relevant problems contain one hundred images at best. Nowadays it becomes necessary to be able to work with photos of documents of a bad quality due to the prevalence of smartphones and digitalization of the document exchange process.

The absence of large-scale document image datasets is a serious problem that has an impact on the current state of research in this area. It raises the entry barrier by throwing out researchers who do not have the resources to create their own datasets. Moreover, it is also difficult to compare different models with each other because they are tested on different datasets, which often have a small size. To overcome these difficulties we present the DDI-100 dataset, which is larger than existing datasets. It is based on publicly available documents and reports, extended by various geometric deformations and distortions. We believe that this dataset of document images will push the creation of more advanced models in the field of document image analysis and allow researchers to compare and test different approaches. The dataset is publicly available at https://github.com/machine-intelligence-laboratory/DDI-100.

---

[*] Corresponding author: ilya.zharikov@phystech.edu

The rest of this paper is organized as follows. Firstly, we consider related datasets and compare them with the DDI-100 dataset. In section 3 we provide a detailed description of the DDI-100 in its current state. Section 4 is devoted to experimental baselines that demonstrate performance of the dataset for OCR and TD. Section 5 concludes the results and contribution of this paper.

## 2 Related Datasets

The main contribution of this work is a large-scale dataset for detecting and recognizing text in the field of document images processing. For this reason, we restrict the discussion to related datasets.

The first dataset for text detection and recognition was the ICDAR Robust Reading challenge [9] that is known as ICDAR 03. It consists of 509 scene images with centered text. After the further iterations of the datasets [10-11] that contains only horizontal English texts in [12] authors presented a dataset of 89 images with text of various directions. To overcome the problem of small data size a new dataset MSRA-TD500 [13] was realized. It includes 500 images of indoor and outdoor scenes. Another example of the natural scene text dataset is SVT dataset [14] that harvested from Google Street View. The dataset contains 350 total images and 725 total labeled words, which often has low resolution. However, it comprises annotations not for all text in the images.

The newest iteration of the ICDAR Robust Reading challenge [15] introduced a dataset of 561 images with a minimum size of 100×100 pixels. The authors analyzed 315 Web pages, 22 spam and 75 ham emails and extracted all the images with text. Based on these images the dataset for the word recognition problems was also collected. It consists of 5003 words with a length of at least 3 characters long. In [16] a new challenge on Incidental Scene Text was presented that focuses on real scene images. The dataset includes Latin-scripted text and text in a number of Orient scripts. It contains 1670 images and 17548 annotated regions. The ground truth for this challenge comprises word-level bounding boxes with their Unicode transcriptions making the dataset suitable for text localisation and recognition problems.

One of the largest and public domain datasets is COCO-Text dataset [17] that is based on MS COCO [18] and its image captions extension [19]. It includes 63686 images with 173589 labeled text regions. The ground truth contains bounding boxes, classifications of each box in terms of legibility. The dataset comprises both machine printed and handwritten texts. Another example of a large-scale dataset is called SynthText in the Wild [20]. Unlike all previous datasets, SynthText is synthetic. The authors presented a new method for generating images with text based on various deep learning techniques. This approach is used to generate 800000 scene-text images. The text for the images was extracted from the Newsgroup20 dataset [21]. To ensure proper diversity 8000 background images are extracted from Google Image Search related to different queries.

The latest presented dataset is called FUNSD [22]. FUNSD is based on a subset of the RVL-CDIP dataset [23] that contains grayscale images of various documents from 80's-90's. The introduced dataset consists of randomly sampled 200 images with more than 30000 word-level annotations.

The key differences of the presented dataset DDI-100 are the following. Fist, DDI-100 has a much larger scale than other many datasets and can be easily extended by adding documents in the appropriate format. Second, all publicly available datasets are useful in the tasks related to text detection and recognition, but they do not reflect the specifics of the document processing research area. Third, DDI-100 contains a wide variety of real multilingual text instances. Figure 1 gives an overview of the datasets in terms of size and number annotated text regions.

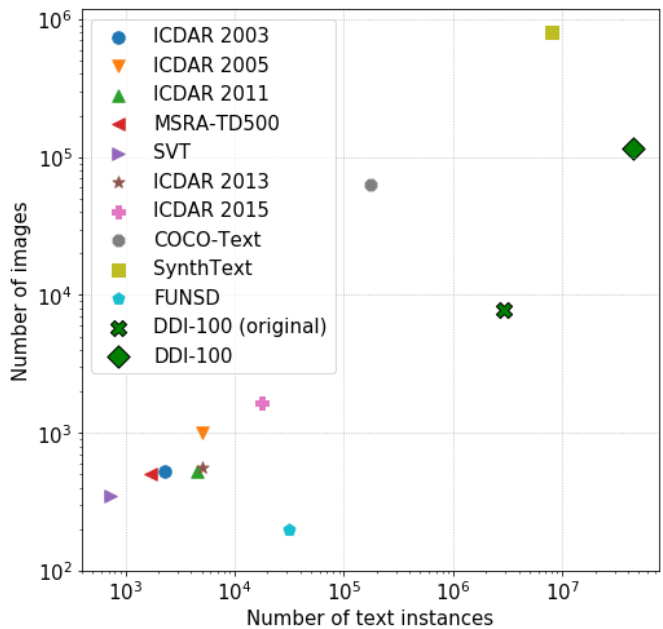

**Fig. 1.** Different statistics of datasets for text detection and recognition.

## 3 Dataset Structure

### 3.1 Image data description

The DDI-100 dataset contains more than 100000 distorted images of 7351 unique documents pages. All documents are in the public domain and include various reports, books, etc. During the generation process 5659 different images were used as backgrounds and textures as well as 99 stamp images. From each unique document page, we have collected 15 different images by applying various types of distortions and geometric transformations (Figure 2). The list of all distortions is the following:
- perspective transformations;
- background replacement;
- document displacement;
- texture mapping;
- text and background overlay with various alpha channel;
- Gaussian and motion blur;
- adding color gradient;
- adding glares and shadows;
- image rescaling;

- stamp overlaying.

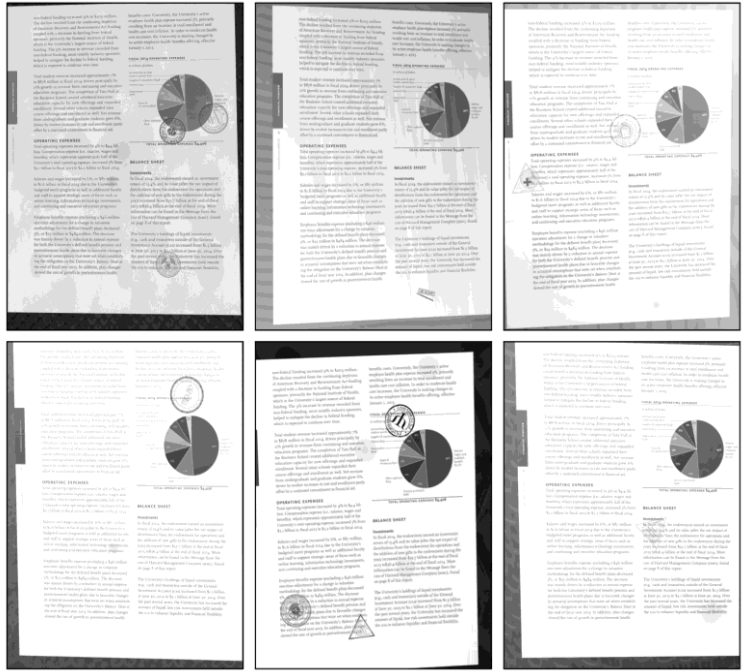

**Fig. 2.** Examples of various distortions applied to the image.

Dataset is divided into 38 unequal parts. Every part corresponds to a single book or report. Each folder contains the following information:
- original file in pdf format;
- original backgrounds as pdf file and as set of images;
- original masks as pdf file and as set of grayscale images;
- text masks as pdf file and as set of grayscale images;
- text blocks positions as set of pickle files;
- generated images;
- text blocks positions of generated images as set of pickle files;
- text masks of generated images;
- stamp masks of generated images.

## 3.2 Ground truth

According to the structure of the DDI-100 dataset described above for each image we have prepared stamps (Figure 3) and text masks (Figure 4).
For text detection and recognition tasks, we have also prepared text annotation information with text values including letter separation information. The ground truth is given in pickle format (Figure 5).

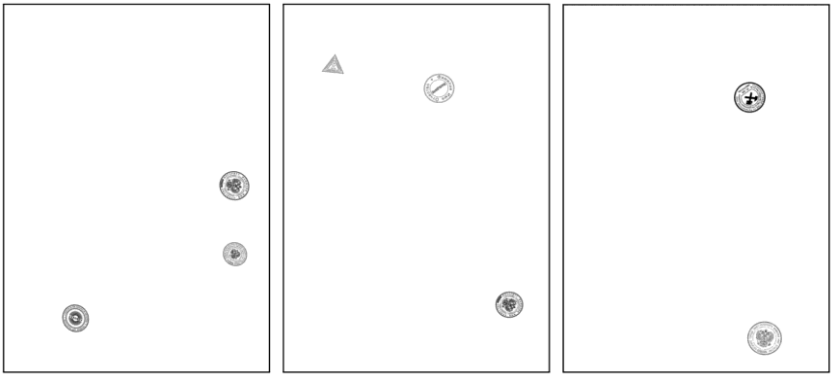

**Fig. 3.** Examples of stamp masks.

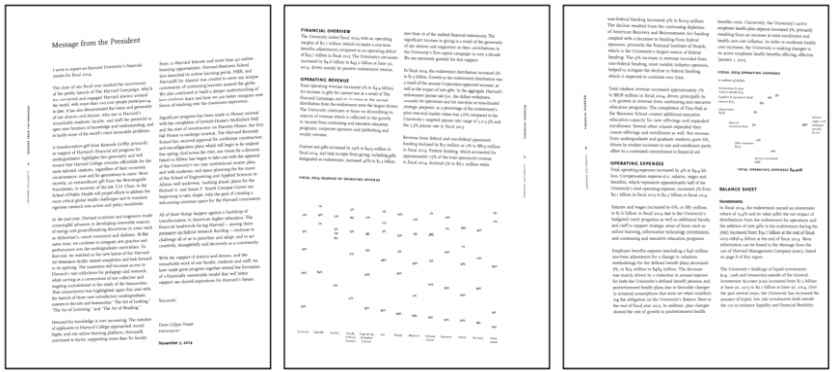

**Fig. 4.** Examples of text masks.

```
{
    'text': 'effective',
    'box': array([[1569, 1906], [1538, 1906], [1569, 2052], [1538, 2052]]),
    'chars': [
        {
            'text': 'e',
            'box': array([[1569, 1906], [1548, 1906], [1569, 1923], [1548, 1923]])
        },
        {
            'text': 'f',
            'box': array([[1569, 1926], [1538, 1926], [1569, 1938], [1538, 1938]])
        },
        {
            'text': 'f',
            'box': array([[1569, 1939], [1538, 1939], [1569, 1951], [1538, 1951]])
        },
        {
            'text': 'e',
            'box': array([[1569, 1954], [1548, 1954], [1569, 1971], [1548, 1971]])
        },
        {
            'text': 'c',
            'box': array([[1569, 1975], [1548, 1975], [1569, 1990], [1548, 1990]])
        },
        ...
    ]
}
```

**Fig. 5.** Image ground truth pickle format for text detection and recognition.

### 3.3 Dataset split

The dataset is split into training and validation set, which contain 70% and 30% images respectively. In each folder, we have fixed the division to ensure that each book is presented in the same way in the training and validation sets.

### 3.4 DDI-100 applications

The dataset is devoted to the following scientific research challenges in document images analysis:
- Text detection. For each image, ground-truth includes text positions in terms of bounding boxes.
- Optical character recognition. For each image, ground-truth contains annotations for all text in the images and letters position for each annotation.
- Stamp detection. Dataset includes images with stamps and ground-truth contains corresponding masks.

## 4 Experiments and results

### 4.1 Text Detection

We present baseline results for TD and OCR systems trained on DDI-100 dataset to show its quality. Moreover, we compare our dataset with the FUNSD and Real-DDI datasets using fixed models. Real-DDI is a dataset which was obtained by the following process. We print 100 unique pages from DDI-100 and take photos using different smartphones on a flat surface under different lighting condition. The pages are manually labelled: bounding boxes of words and their text annotations. The dataset is divided into two parts: 20 photos for test and 80 photos for train phase.

The TD on the DDI-100 is tested with three baselines: an efficient and accurate scene text detector(EAST)[24], connection text proposals network(CTPN)[25] and U-net[26]. EAST and CTPN are the popular open-source solutions, we use them without re-training on the DDI-100. As a third baseline, we use a two-step model: the first part is a neural network that predicts a mask where every pixel represents the probability of belonging to the 'text' class. In the second step, the outputs are converted into bounding boxes.

For the first step, we use U-net architecture with four encoder and decoder blocks. Skip connection[26] is used as a relation between encoder and decoder. Each block consists of a serial application of convolution, batch normalization, ReLU nonlinearity, convolution, batch normalization, ReLU nonlinearity. Maxpooling reduces dimension in the encoder, upsampling rises in the decoder. All images are resized to 1920x1280 resolution by scaling and padding. The loss function is a linear combination of IOU and binary cross-entropy. We use erosion-dilatation with a small kernel to remove noise in the network output. The resulting connected areas are approximated by rotated rectangles.

To evaluate the model performance, we compute such metrics as precision, recall, f-measure. Box are considered as correctly predicted if IOU metric is above 0.8. Results can be seen in Table 1.

**Table 1.** Performance of text detection baseline (P – precision, R – recall, F – f-score).

|  | FUNSD | | | Real-DDI | | | DDI-100 | | |
|---|---|---|---|---|---|---|---|---|---|
|  | P | R | F | P | R | F | P | R | F |
| U-net | 0,857 | 0,917 | 0,886 | 0,974 | 0,983 | 0,978 | 0,932 | 0,964 | 0,948 |
| EAST | 0,961 | 0,591 | 0,731 | 0,984 | 0,887 | 0,933 | 0,962 | 0,853 | 0,905 |
| CTPN | 0,833 | 0,601 | 0,699 | 0,878 | 0,862 | 0,87 | 0,797 | 0,762 | 0,779 |

According to the results, our U-net based model trained on the DDI-100 dataset achieve better quality compared to popular solutions for text detection problem.

## 4.2 OCR

In order to test quality of DDI-100 for training OCR systems two baseline models were used: Tesseract and Neural machine translation (NMT). Tesseract model was tested without re-training on the dataset, it was chosen as most common open-source solution for OCR systems. We use our own implementation of NMT with following architecture. The model takes images with a fixed height and varying width. It includes two main parts. The first one is convolution neural network that creates vector representation for the parts of the source image. In the second part, these vector representations are used as an input for NMT. We apply gated recurrent units(GRU) for encoder and decoder with Bahdanau attention. The second part produces embeddings of symbols which are located on the source image. This end-to-end model is trained with weighted cross-entropy loss.

To evaluate the solutions, we use the following metrics: word match accuracy and normalized Levenshtein distance. We observe that Tesseract returns an empty string on the part of the data, so we decided to evaluate metrics among not empty predictions too. The result of the model is presented in the Tables 2 and 3.

**Table 2.** Part of empty predictions for Tesseract.

|  | Real-DDI | DDI-100 | FUNSD |
|---|---|---|---|
| Empty prediction, % | 24 | 50 | 56 |

**Table 3.** Results for Tesseract model (WMA – word match accuracy, NLD – normalized Levenshtein distance).

|  | Real-DDI | | DDI-100 | | FUNSD | |
|---|---|---|---|---|---|---|
|  | WMA | NLD | WMA | NLD | WMA | NLD |
| Tesseract(not empty) | 0.26 | 0.39 | 0.2 | 0.36 | 0.06 | 0.54 |
| Tesseract(full) | 0.19 | 0.54 | 0.1 | 0.68 | 0.03 | 0.8 |

In the next experiment, NMT model was trained on a dataset (DDI-100, real-DDI, FUNSD) and evaluated on test part of each dataset. Model trained on FUNSD was evaluated on part of DDI-100 and real-DDI boxes which contains English text. From Table 4, we observe that NMT model outperforms Tesseract. Moreover, our dataset helps to solve the OCR problem on a satisfactory level, while avoiding a costly procedure of labelling data.

**Table 4.** Performance of NMT OCR baseline (WMA – word match accuracy, NLD – normalized Levenshtein distance).

|  | Real-DDI | | DDI-100 | | FUNSD | |
| --- | --- | --- | --- | --- | --- | --- |
|  | WMA | NLD | WMA | NLD | WMA | NLD |
| NMT(DDI-100) | 0.67 | 0.13 | 0.85 | 0.04 | 0.10 | 0.66 |
| NMT(FUNSD) | 0.42 | 0.26 | 0.35 | 0.33 | 0.65 | 0.13 |
| NMT(Real-DDI) | 0.74 | 0.096 | 0.5 | 0.26 | 0.14 | 0.60 |

The last experiment is devoted to demonstrate the advantages of model pre-trained on the DDI-100 dataset. We take pre-trained on the DDI-100 NMT model and train it on real-DDI and FUNSD. It clearly comes from Table 5 that extra training on a specific dataset helps to improve the model performance to match the state-of-the-art solutions. While we need 15000 iterations of gradient descent to achieve local minimum (batch size is 64) from random initialization, 600 iteration of extra training is enough to achieve high quality solution.

**Table 5.** Performance of extra training of NMT pre-trained on DDI-100 (WMA – word match accuracy, NLD – normalized Levenshtein distance).

|  | Real-DDI | | FUNSD | |
| --- | --- | --- | --- | --- |
|  | WMA | NLD | WMA | WMA |
| NMT(FUNSD) |  |  | 0.69 | 0.09 |
| NMT(real-DDI) | 0.87 | 0.04 |  |  |

## Discussion

The main contribution of this work is the DDI-100 dataset for text detection and recognition in document images. This dataset is based on publicly available reports, papers, books, etc. It is the first large-scale dataset in the field of document images processing containing document-specific features like stamps, tables, diagrams and dense text. DDI-100 is a semi-synthetic dataset with real textual content that provides detailed ground truth annotations which are cheap and scalable by comparison to real data. Obtained experiment results demonstrate the usefulness of DDI-100 that allows to achieve the state-of-art solutions using a small number of real annotated instances. This motivates future work on expanding dataset possible applications by adding new languages, symbols and distortions including nonlinear transformations.